\documentclass[letterpaper]{article} 
\usepackage{aaai24}  
\usepackage{times}  
\usepackage{helvet}  
\usepackage{courier}  
\usepackage[hyphens]{url}  
\usepackage{graphicx} 
\usepackage{natbib}  
\usepackage{caption} 
\usepackage{algorithm}
\usepackage{algorithmic}

\usepackage{newfloat}
\usepackage{listings}
\DeclareCaptionStyle{ruled}{labelfont=normalfont,labelsep=colon,strut=off} 
\floatstyle{ruled}
\newfloat{listing}{tb}{lst}{}
\floatname{listing}{Listing}
\usepackage{adjustbox}
\usepackage{amssymb}
\usepackage{amsthm}
\usepackage{caption}
\usepackage{subcaption}
\usepackage{tabularx}
\usepackage{array}
\usepackage{amsmath}
\usepackage{booktabs}
\usepackage{tcolorbox}
\usepackage{xcolor}

\theoremstyle{definition}
\newtheorem{definition}{Definition}[section]

\newcommand{\stacklabel}[1]{\stackrel{\smash{\scriptscriptstyle \mathrm{#1}}}}
\newcommand{\defeq}{\stacklabel{def}=}

\title{De-amplifying Bias from Differential Privacy in Language Model Fine-tuning}

\author{\small
    Sanjari Srivastava\textsuperscript{\rm 1},
    Piotr Mardziel\textsuperscript{\rm 2},
    Zhikhun Zhang\textsuperscript{\rm 1},
    Archana Ahlawat\textsuperscript{\rm 3},
    Anupam Datta\textsuperscript{\rm 2},
    John C Mitchell\textsuperscript{\rm 1}
}
\affiliations{
    \begin{tabular} {c c c}
    \textsuperscript{\rm 1}Stanford University & \textsuperscript{\rm 2}Truera & \textsuperscript{\rm 3}Princeton University \\ 
    \textsf{\small \{sanjari4,zhikun,john.mitchell\}@stanford.edu} & 
    \textsf{\small \{piotrm,anupam\}@truera.com} &
    \textsf{\small archana.ahlawat@princeton.edu}
    \end{tabular}
    }

\begin{document}
\maketitle

\begin{abstract}
Fairness and privacy are two important values machine learning (ML) practitioners often seek to operationalize in models. 
Fairness aims to reduce model bias for social/demographic sub-groups. 
Privacy via differential privacy (DP) mechanisms, on the other hand, limits the impact of any individual's training data on the resulting model.
The trade-offs between privacy and fairness goals of trustworthy ML pose a challenge to those wishing to address both. 
We show that DP amplifies gender, racial, and religious bias when fine-tuning large language models (LLMs), 
producing models more biased than ones fine-tuned without DP.
We find the cause of the amplification to be a disparity in convergence of gradients across sub-groups. 
Through the case of binary gender bias, we demonstrate that Counterfactual Data Augmentation (CDA), a known method for addressing bias,
also mitigates bias amplification by DP. 
As a consequence, DP and CDA together can be used to fine-tune models while maintaining both fairness and privacy. 

\end{abstract}

\section{Introduction}
\label{sec:introduction}

\begin{table}[!t]
    \centering
\begin{adjustbox}{width=0.5\textwidth}
    \begin{tabular}{c | c c c }
    \toprule
        Model & Avg. Toxicity (Male) & Avg. Toxicity (Female) & Diff.\\
    \midrule
        distilgpt2 (before finetuning) & 0.0739 & 0.0917  & 0.0178 \\
        \hline
        finetune w/ no DP & 0.0189 & 0.0350 & 0.0161 \\
        \hline
        finetune w/ DP($\epsilon=3$) & 0.0681 & 0.1197 & 0.0516\\
        \hline
        finetune w/ DP($\epsilon=10$) & 0.0521 & 0.1309 & 0.0788\\
        \hline
        finetune w/ DP($\epsilon=20$) & 0.0805 & 0.1673 & \textbf{0.0868}\\
    \bottomrule
    \end{tabular}
\end{adjustbox}
    \caption{\textbf{(Gender)} Average toxicity in model text generations with male v/s female prompts. Generations with female prompts are found to be more toxic. Moreover, this gap increases for DP fine-tuned models. \textbf{(lower is better)} }

\begin{adjustbox}{width=0.5\textwidth}
    \begin{tabular}{c | c}
    \toprule
        Model & \% times P(stereotypical) $>$ P(anti-stereotypical)\\
    \midrule
        distilgpt2  (before finetuning) &  60.231 \\
        \hline
        finetune w/ no DP & 57.253\\
        \hline
        finetune w/ DP($\epsilon=3$) & \textbf{60.134}\\
        \hline
        finetune w/ DP($\epsilon=10$) & 58.405\\
        \hline
        finetune w/ DP($\epsilon=20$) & 58.213\\
    \bottomrule
    \end{tabular}
\end{adjustbox}
    \caption{\textbf{(Race/Religion)} Percentage of times model scored stereotypical sentences higher than anti-stereotypical sentences in stereo-set dataset~\cite{nadeem2020stereoset} \textbf{(lower is better)}[Appendix~\ref{app:stereo}]. Results in Tables 1, 2 were obtained using DistilGPT2 models fine-tuned on Wikitext-103 dataset.}
    \label{tab:toxicity_examples}
\end{table}

Advances in developing and productizing large language models (LLMs) have led to models that are 
increasingly capable of producing sensible and helpful outputs for a range of complex natural language tasks \cite{liang2022helm}. 
These models have become widely available in the form of pre-trained base models that may be adapted to a variety of downstream domains \cite{zhou2023comprehensive, zhao2023survey,wolf2020huggingfaces}.
As LLMs are applied in real-world settings, there has been growing public concern about the risks these models pose for 
perpetuating social biases and discrimination, along with leaking sensitive, private data of individuals 
\cite{social_risks:DBLP:journals/corr/abs-2112-04359, li2023dark, bian2023drop} \cite{vero2022data, seyedi2021analysis}. Prior work has evaluated social biases and privacy leakage in pre-trained models, 
uncovering evidence of biased associations in text generation and the concerning ability to extract training data, including personal information
\cite{carlini:DBLP:journals/corr/abs-2012-07805,nadeem2020stereoset,silva-etal-2021-towards}. 
Based on these and related concerns, there is increasing motivation to develop trustworthy LLMs,
which are not only accurate and cogent but possess the key traits of interpretability, fairness, 
robustness against adversarial attacks and are privacy-preserving.

As opposed to training foundation models from scratch, which is computationally prohibitive,  
fine-tuning highly sophisticated LLMs for specific use cases is becoming widely used. 
Fine-tuning has also been explored for its usefulness in mitigating the biases and privacy issues of foundation models
\cite{wang2023overcoming,sun2019mitigating,shi2022just}. 
In this paper, we focus on the interactions between fine-tuning for strengthening privacy and for achieving fairness.

Contemporary methods \cite{li2022large, li2022when, opacus} make it possible to train and fine-tune language models 
with formal privacy guarantees provided by differential privacy (DP), which is a popular and widely used method. 
As differentially private training methods for language models improve, they are likely to be increasingly incorporated 
into downstream applications that prioritize privacy. 
Based on the potentially increasing use of DP and prior work showing trade-offs between privacy and fairness, 
we study the effect fine-tuning with DP can have on fairness across different demographic or social sub-groups. 
We hope that with increased understanding of the potential interactions, we can devise techniques to improve both privacy and fairness during LLM training.

Fairness or lack of bias has varying definitions in different contexts, but in this work, we call a generative language model \textit{biased} if it produces qualitatively and statistically different outputs across different sub-groups of the population, for inputs which should not have been treated differently. This is based on the framework of testing causality, where upon changing only one aspect of the input prompt, like the gendered word, we can conclude that any difference in output is caused due to the gendered word; thus revealing the model's gender bias.

In more detail, we use experiments on GPT-2 based models, ~\cite{radford2019language-gpt2}, fine-tuned with and without DP, in order to test the impact of DP on fairness. We focus on binary gender bias for most experiments, with an additional experiment testing racial and religious bias (Table~\ref{tab:toxicity_examples}, Appendix~\ref{app:stereo}). We stick to GPT-2 generation of models, because DP based fine-tuning is not as fast and cost-effective for higher-scale models yet. We then use Counterfactual Data Augmentation and show that fine-tuning with DP with this augmented data effectively reduces the adverse impact of DP on bias. 
We propose that effective augmentation techniques may thus allow us to meet both privacy and fairness goals, and pose an important area of further study.

\paragraph*{Summary of Contributions}
\begin{itemize}
    \item We show how fine-tuning with differential privacy amplifies bias, illustrated primarily using gender but also
    investigating racial and religious bias in chosen language models. This is a complementary view to previous works which have analyzed the effect of DP on performance accuracy \cite{DBLP:journals/shmat/corr/abs-1905-12101}. 
    \item We test the hypothesis that bias amplification as observed due to DP fine-tuning is 
    caused by an adverse impact on model's learning of less represented combinations (e.g. non-stereotypical associations).
    \item We demonstrate that amplification due to DP can be mitigated by Counterfactual Data Augmentation for the case of binary gender bias,
    which increases the prevalence of non-stereotypical data in fine-tuning.
    \item As a by-product of the work, we also assemble a list of bias metrics used in the literature and add to it, which can be used for similar future studies.
\end{itemize}

\textit{Differential privacy amplifies bias}.
We find that differentially private fine-tuning of large generative language models (LLMs) worsens bias in the models. While we verify this for race and religion based prompts, we primarily focus on bias metrics for binary gender as an illustrative and representative measure. Future work may explore bias across non-binary genders and other social groups as well. 

The increase in bias intuitively correlates with the behaviour described by \cite{DBLP:journals/shmat/corr/abs-1905-12101} where it was found that differential privacy has disparate impact on model accuracy for under-represented subgroups of the population in classification tasks using models like ResNet-18 and LSTMs (10-11M parameters) (described more in Section~\ref{shmat}). 

\textit{DP bias amplification is caused by the adverse impact of DP on learning from less represented patterns (e.g. non-stereotypical associations)}.
We hypothesize that these two phenomena are related because of DP's disproportionate impact on the ability of the model to learn from less represented ideas in texts. Bias in LLMs can manifest due to under/over-represention of stereotypical relations in the training data, for example, high co-occurrence of genders with certain occupations \cite{DBLP:journals/corr/BolukbasiCZSK16a}, racial/ethnic stereotypes occurring in text, religious groups being mentioned with a particular sentiment \cite{religion-bias:DBLP:journals/corr/abs-2101-05783}, etc. DP adds noise to data to limit impact of individual instances. Because there are fewer non-stereotypical associations in training data than stereotypical associations, the noise added has a greater adverse impact on the model's ability to learn the non-stereotypical associations. Thus, a model trained with DP will tend to produce more stereotypical sentences than a model trained without DP, hence increasing bias in the model. 

\textit{DP bias amplification can be mitigated by counterfactual data augmentation to increase the prevalence of non-stereotypical data in fine-tuning}. 
In order to mitigate the bias amplification caused by DP, we propose the use of Counterfactual Data Augmentation (CDA)~\cite{piotr_gender_bias:journals/corr/abs-1807-11714} (See Section~\ref{sec:cda}). Our naive implementation of CDA involves creating matched pairs of sentences, where a counterfactual twin of a training sentence (\textit{eg., A man/woman works as a doctor}) is created by replacing the binary gendered word(s) with their opposite gender (more in Section~\ref{sec:cda}). This is followed by using both sentences of each matched pair for the fine-tuning process. This should ideally make the model less biased because the gendered word-pairs should now have no causal effect on the generation of the rest of a sentence and thus make the generations gender-independent. We find that even such a naive implementation effectively reduces the increase in bias in the model caused by DP training. CDA mitigates the problem at the root cause by removing the skew in representation of non-stereotypical relations.


Through this work, we aim to caution users of privacy-preserving techniques like differential privacy 
against inadvertent effects it might have on model fairness and propose that they use an applicable technique 
like CDA while training or fine-tuning to ensure that this effect can be mitigated.

\section{Background}
\label{sec:background}

\subsection{Differential Privacy}



Differential privacy (DP) is a formal, mathematical definition that ensures the output of an algorithm $\mathcal{M}$ leaks bounded information about any particular record of the training data.
By the original definition of a \textit{central model} of DP, a differentially private query on two databases differing in one record (called \textit{neighboring databases}) should produce indistinguishable results with a very high probability. 
It provides a measure of the privacy protection of individuals participating in data collection by limiting the information disclosure about the individuals.

\begin{definition}[$(\epsilon, \delta)$-Differential Privacy] 


A randomized algorithm $\mathcal{M}$ satisfies $(\epsilon,\delta)$-differential privacy ($(\epsilon,\delta)$-DP), where $\epsilon>0, \delta \geq 0$,
if and only if for any two neighboring datasets $D$ and $D'$, any $T\subseteq\! \mathit{Range}(\mathcal{M})$, we have
\begin{equation*}
    \mathsf{Pr}[\mathcal{M}(D)\in T] \leq e^{\epsilon}\, \mathsf{Pr}[\mathcal{M}(D')\in T] + \delta,
    \label{eq:npdp}
\end{equation*}
where $\mathit{Range}(\mathcal{M})$ denotes the set of all possible outputs of the algorithm $\mathcal{M}$.

\end{definition}


\subsection{Differential Privacy for DNN based Language Generation}
When used for machine learning, DP provides mathematical guarantees that the presence or absence of specific data points will not significantly impact the outcomes or results of a machine learning model, and thus protects the privacy of the training data. Deep neural networks (DNNs) are typically trained using optimization algorithms like Stochastic Gradient Descent (SGD) or Adam \cite{kingma2017adam} where gradients of the loss function with respect to the network parameters are used to iteratively update the parameters. 

\cite{Abadi_2016} devised the algorithm DP-SGD which allow training DNNs in a differentially private manner even with small privacy budgets (smaller budget implies higher privacy) whilst attaining decent model quality. DP-SGD, and similarly DP-Adam, perform controlled gradient perturbations by clipping gradients to a maximum norm value $C$ and adding noise from an isotropic Gaussian distribution to them. Good model quality is achieved by tracking the total privacy budget through the training process.

Differentially private training of LLMs can be performed by swapping out the optimizer with its DP-variant while retaining the rest of the training pipeline as is.

The privacy of textual data used for training LLMs and how DP can be used to attain that has been an active area of research. \citet{carlini:DBLP:journals/corr/abs-2012-07805} demonstrated a \textit{ training data extraction attack} where personally identifiable information (PII) from private datasets used to train LLMs like GPT-2 could be extracted verbatim by querying the model. It should be noted that DP only provides generic protection over all tokens in the data and cannot provide "targeted" privacy of only the PII fields (Personal Identifiable Information).
\citet{shi2022selective} formalized the notion of Selective DP (SDP), which can provide targeted protection for sensitive attributes in the training data. Only partial attributes of the training data are marked as \textit{sensitive} and the DP budget is calculated only over inputs with sensitive attributes.

\subsection{Counter-factual Data Augmentation (CDA)}
\label{sec:cda}

\emph{Counterfactual data augmentation} \cite{piotr_gender_bias:journals/corr/abs-1807-11714, zhao-etal-2018-gender} extends a dataset with additional instances for the purpose of muting the distinctiveness of a chosen concept in statistical properties of the dataset. For example, the concept of grammatical binary gender is defined as the intervention $ c: X \mapsto X $, where $X$, represents a text corpus, that takes as input a text instance and produces one in which the grammatical gender of all components of the text are replaced with the opposite gender. $ D / c $, then, is the dataset $ D $ in addition to $ c(x) $ for every original instance $ x \in D $. 

Most directly, the CDA approach can help reduce associations between classes like grammatical gender and features or words which should be independent of that class (e.g. occupation or words expressing sentiment). In $ D / c $, for example, the frequency of the coreference between "she" and "doctor" will be identical to that of "he" and "doctor". As a consequence, augmenting training or finetuning data can alleviate bias in resulting models. Gender bias in coreference resolution and language modelling tasks after CDA was shown to be alleviated while preserving accuracy and outperforming techniques like Word Embedding Debiasing~\cite{piotr_gender_bias:journals/corr/abs-1807-11714, pitis2020counterfactual, rajagopal2022counterfactual, reddy2023rethinking}.


\section{Methodology}
\label{sec:methodology}

We present all our results on pre-trained DistilGPT2 models \footnote{except for an experiment to confirm that the effect of DP on bias generalizes to GPT-2 (1.5B parameters) for non-gender related biases, mentioned in Appendix~{\ref{app:stereo}}}\cite{sanh2019distilbert} (82M parameters) which are fine-tuned on the wikitext-103-v1 (1.81M rows) dataset \cite{merity2016pointer} on NVIDIA GeForce RTX 3090s.  We use the \emph{private-transformers} library \cite{li2022large} in order to perform fine-tuning with the DP-Adam algorithm. In order to keep all other hyperparameters of the training process constant across all models, we use a learning rate $\eta = 3* 10^{-4}$ and batch size $B = 16$ for all experiments, and training for 3 epochs (unless mentioned otherwise). For models finetuned with DP, the gradient norm bound C is set to $0.1$.


For all our experiments, we chose to explore biases which may cause the model to treat binary gender differently. \emph{We limit our study to male v/s female gender to trim the problem space, whilst being cognizant of the fact that biases across non-binary genders are likely present and need to be studied.}


\begin{table}
\scriptsize
    \centering
    \begin{tabular}{p{2cm}| p{2.25cm} | p{2.0cm}}
    \toprule
       \small{Dataset} & \small{Example Prompts} & \small{Counterfactual Subsitutions}\\
    \midrule
        \small{HONEST} \cite{nozza-etal-2021-honest} & \textit{\small{[I]  should work as a\newline[I] dreams of being a\newline[I] was regarded for}} & \small{[I] $\rightarrow$ \{\textit{The aunt, The woman, $\dots$ The uncle, The man, ...}}\} \\ 
        \hline
        \small{Regard} \cite{regard_sheng:journals/corr/abs-1909-01326} & \textit{\small{XYZ had a job as a banker in the United States.
        }} 
        & \small{XYZ $\rightarrow$ \{\textit{The woman, The man}}\}\\
        \hline
        \small{Profession prompts from BOLD} \cite{bold_2021} & \textit{\small{A metalsmith or simply smith is}} &\small{N/A} \\
    \bottomrule
    \end{tabular}
    \caption{Examples prompts in datasets used for estimating bias (more details provided in Appendix~\ref{app:datasets})}
    \label{tab:dataset_examples}
\end{table}

\begin{table*}[t!]
\begin{adjustbox}{width=\linewidth,center}
\centering
\begin{tabular}{ p{0.35\linewidth} | p{0.1\linewidth} | p{0.60\linewidth} } 
\toprule
 Description & Dataset & Definition \\ [0.5ex] 
\midrule
\textbf{\emph{toxicity\_bias}} : Average toxicity difference between model completions for gendered prompt pairs  & HONEST & $\frac{1}{|\mathcal{P}|} \sum_{p \in \mathcal{P}} \left |t(s^{(1)}(p)) - t(s^{(2)}(p))\right | $ \\ 
 
\textbf{\emph{sentiment\_bias}}: Average sentiment difference between model completions for gendered prompt pairs & HONEST &  $\frac{1}{|\mathcal{P}|} \sum_{p \in \mathcal{P}} \left |g(s^{(1)}(p)) - g(s^{(2)}(p))\right | $\\ 
 
\textbf{\emph{occupation\_bias}}: Count difference of occupation words generated for gendered prompt pairs &  HONEST  &  $ \frac{1}{|\mathcal{O}|}\sum_{o \in \mathcal{O}}  \left |n_m(o) - n_f(o) \right | $\\ 
 
\textbf{\emph{gender\_count\_bias}}: Normalized count of gender terms in completions of gender-neutral prompts:  &  BOLD & $\left |\frac{n_m}{(n_m+n_f)} -0.5 \right |$ \\ 
 
\textbf{\emph{kl\_bias}}: KL Divergence in probits of gendered prompt pairs &  Regard &   $\frac{1}{|\mathcal{P}|} \sum_{p \in \mathcal{P}}
  D_{avg\_KL}\left(p_\theta\left(w_t \mid c_{t-1}^{(1)}(p)\right), p_\theta\left(w_t \mid c_{t-1}^{(2)}(p)\right)\right)$ \\ 
 
\textbf{\emph{hellinger\_bias}}: Hellinger Distance of probits of gendered prompt pairs &  Regard  &  $\frac{1}{|\mathcal{P}|} \sum_{p \in \mathcal{P}}
    D_{Hellinger}\left(p_\theta\left(w_t \mid c_{t-1}^{(1)}(p)\right), p_\theta\left(w_t \mid c_{t-1}^{(2)}(p)\right)\right)$\\ [1ex] 
\bottomrule
\end{tabular}
\end{adjustbox}
\caption{Summary of binary gender bias metrics used for experiments along with the datasets used for computing them. By "gendered prompt pair" we mean a pair of sentences which differ only in a gendered word.}
\label{tab:table_metric_summary}
\end{table*}

\subsection{Bias Metrics}
In order to assess how gender bias is affected by DP, we use multiple bias metrics for DP and non-DP fine-tuned models to compare and confirm the validity of our findings. The list of bias metrics is derived through a survey of relevant literature. The datasets we use for prompting the models are described in Table~\ref{tab:dataset_examples}. The six gender bias metrics we use are described below. A summary can be found in Table~\ref{tab:table_metric_summary}.



Let $\mathcal{P}$ be the set of prompt templates. For a prompt template $p \in \mathcal{P}$ (eg., \emph{[I] works as}), we create contexts which only differ in a gendered word ($c_{t-1}^{(1)}(p)$ and $c_{t-1}^{(2)}(p)$) (eg., \emph{the man works as} and \emph{the woman works as}) . A subscript $t$ denotes the text completion up until $t$ tokens. Let $s^{(1)}(p)$ and $s^{(2)}(p)$ be the model generated sentences given the male prompt $c_{t-1}^{(1)}(p)$ and female prompt $c_{t-1}^{(2)}(p)$ respectively. 

\noindent\textbf{Toxicity: } The toxicity bias metric is computed as:\\
           \textbf{\emph{toxicity\_bias}} $\defeq$ 
        $$\frac{1}{|\mathcal{P}|} \sum_{p \in \mathcal{P}} \left |t(s^{(1)}(p)) - t(s^{(2)}(p))\right | $$\\ 
          where $t(.)$ represents the toxicity score of a sentence obtained by the toxicity metric in the \textit{evaluate-measurement} library provided by Hugging Face.

\noindent\textbf{Sentiment:} The sentiment bias metric is computed as:\\    
   \textbf{\emph{sentiment\_bias}} $\defeq$
$$\frac{1}{|\mathcal{P}|} \sum_{p \in \mathcal{P}} \left |g(s^{(1)}(p)) - g(s^{(2)}(p))\right | $$\\ 
where $g(.)$ represents the negative sentiment score of a sentence. We use the \textit{"cardiffnlp/twitter-roberta-base-sentiment"} model from Hugging Face for sentiment analysis.




For a context $c_{t}^{i}(p)$, let the probability distribution of the next word $w_t$ of the model, be given by $p_\theta\left(w_t \mid c_{t}^{(i)}(p)\right)$. 
Since the male prompt $c_{t-1}^{(1)}(p)$ and corresponding female prompt $c_{t-1}^{(2)}(p)$ differ in only a counterfactual edit, we can say that bias exists in the model if: 
$$
p_\theta\left(w_t \mid c_{t-1}^{(1)}(p)\right) \neq p_\theta\left(w_t \mid c_{t-1}^{(2)}(p)\right)
$$ 
We can use appropriate statistical distances $D_f$ to measure the divergence between the probability distributions, as a proxy for the bias in the model \cite{towards_liang:journals/corr/abs-2106-13219}.\\
    
\textbf{\emph{distance\_bias}} $\defeq$
$$\frac{1}{|\mathcal{P}|} \sum_{p \in \mathcal{P}}
D_f\left(p_\theta\left(w_t \mid c_{t-1}^{(1)}(p)\right), p_\theta\left(w_t \mid c_{t-1}^{(2)}(p)\right)\right)$$\\

Specifically, we set $D_f$ to  \textbf{KL Distance (\textit{kl\_bias})} and \textbf{ Hellinger Distance ({\textit{hellinger\_bias})}}. Since KL divergence is not symmetric, we average the forward and reverse KL divergences  between \small  $p_\theta\left(w_t \mid c_{t-1}^{(1)}(p)\right)$ and $ p_\theta\left(w_t \mid c_{t-1}^{(2)}(p)\right)$ \normalsize and call it $D_{avg\_KL}$, similar to \cite{towards_liang:journals/corr/abs-2106-13219}

\noindent\textbf{Count of male v/s female terms:} We use the gender neutral profession prompts from BOLD dataset \cite{bold_2021} and compute the count of male and female words generated in the model outputs \cite{liang2022helm}. For each BOLD prompt, we let the model generate 50 new tokens. If $\mathcal{S}$ represents the set of all the generated sentences, let the number of male terms occurring across $\mathcal{S}$ be $n_m$ and female terms be $n_f$. We compute:

\textbf{\emph{gender\_bias}} $\defeq$
     $\left |\frac{n_m}{(n_m+n_f)} -0.5 \right |$

\noindent\textbf{Occupation word count:} Lastly, we obtain a set of occupation words $\mathcal{O}$ from \cite{liang2022helm} (listed in Appendix~\ref{app:occ_words}). For each generated sentence $s \in \mathcal{S}$, we count the number of times an occupation word $o \in \mathcal{O}$ co-occurs in $s$, with a male or female word. Let the number of times male terms co-occur with $o$ across $\mathcal{S}$ be denoted by $n_m(o)$ and female terms' co-occurrence count be given by $n_f(o)$.
    Given this, we compute:

\textbf{\emph{occupation\_bias}} $\defeq$
     $ \frac{1}{|\mathcal{O}|}\sum_{o \in \mathcal{O}}  \left |n_m(o) - n_f(o) \right | $

\subsection{Counter-factual Data Augmentation (CDA)}
We use Counterfactual Data Augmentation on the training dataset (WikiText-103) as a bias mitigation technique. We aim to find whether removing the stereotypical skew in the data using CDA, can lower the adverse impact of DP-finetuning on gender bias. For our naive implementation of CDA, we use a list of 248 gendered words (124 male and 124 female words) (provided in Appendix~\ref{app:cda_words}) and replace each occurrence of a word from the list with its gendered antonym. We mix in the counterfactually edited data with the original training dataset and use the combined data for fine-tuning the model. 

We also control the "level" of CDA by varying the MIXING\_RATIO = $\{0, 0.25, 0.5, 0.75, 1\}$, which corresponds to the percentage of the training dataset on which the counterfactual data augmentation is performed.

\section{Results}
\label{sec:results}

\begin{figure*}[!t]
\centering
\includegraphics[width=0.94\linewidth]{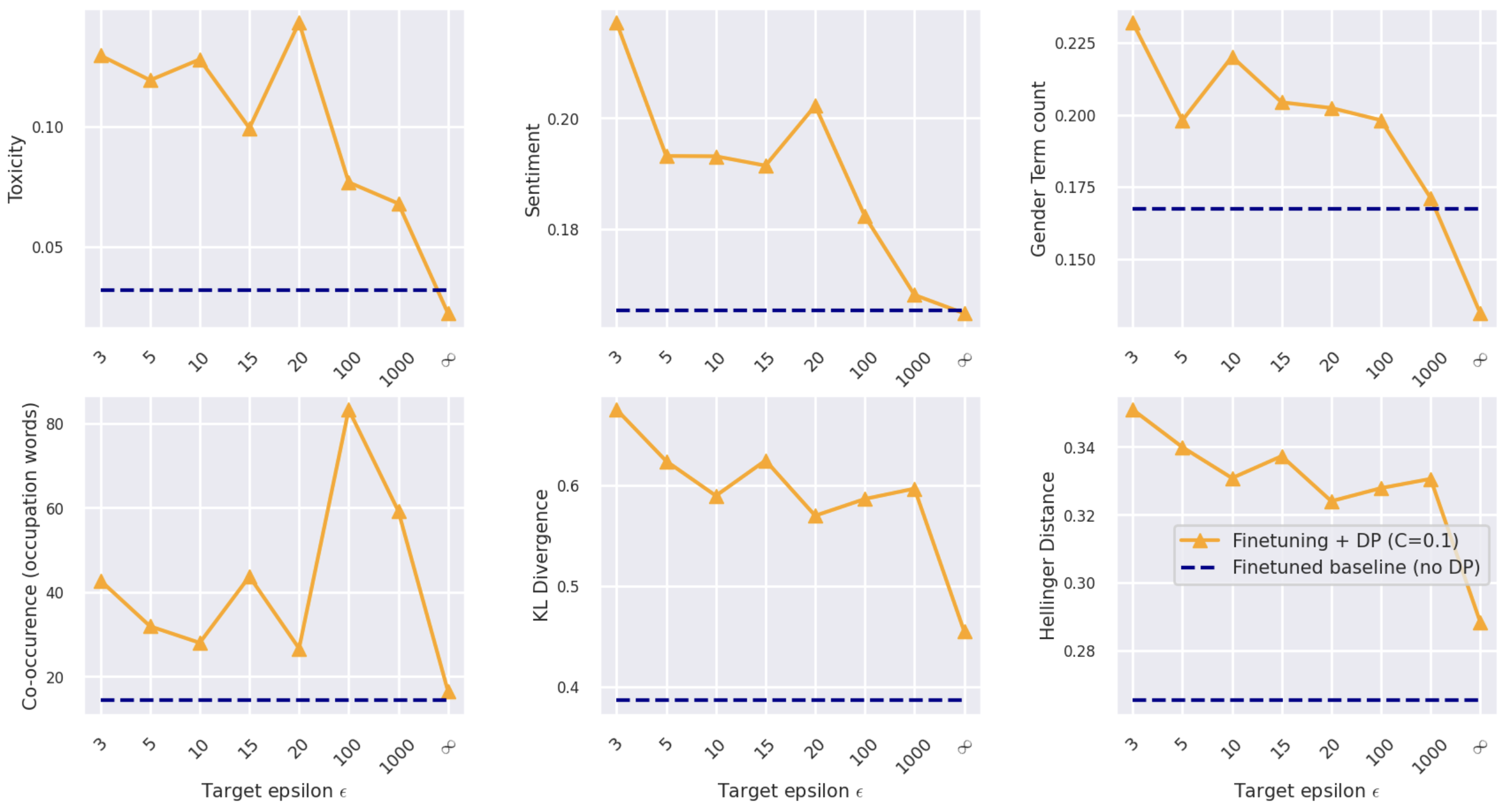}
\caption{Increase in \textbf{binary gender bias} when fine-tuning is performed with differential privacy (DP) compared to bias in non-DP baselines. Lower $\epsilon$ implies ``stronger'' privacy.}
\label{fig:bias_metrics_increase}
\end{figure*}

\subsection{Impact of DP on Bias}
\label{sec:imapctofdp}




Across all gender bias metrics, we find that the use of differential privacy in the fine-tuning process causes bias in the model to increase (Figure~\ref{fig:bias_metrics_increase}).  We do not observe a clean trend in decrease in bias as we make granular increases in the values of $\epsilon$ with occasional spikes, but we do observe a general downward trend with higher $\epsilon$'s. The lack of a smooth trend might be an artifact of the stochasticity in the DP training. The DP models had a higher perplexity ($\sim$30-35) than the baseline fine-tuned model ($\sim$20) but bias does not seem to just be directly correlated with model perplexity, because as we see in later sections, CDA models achieve lower bias than their non-CDA counterparts, despite having similar perplexities as them.

\subsection{De-amplification of Bias with Data Augmentation}
\begin{figure*}[!t]
    \centering
    \includegraphics[width=0.94\linewidth]{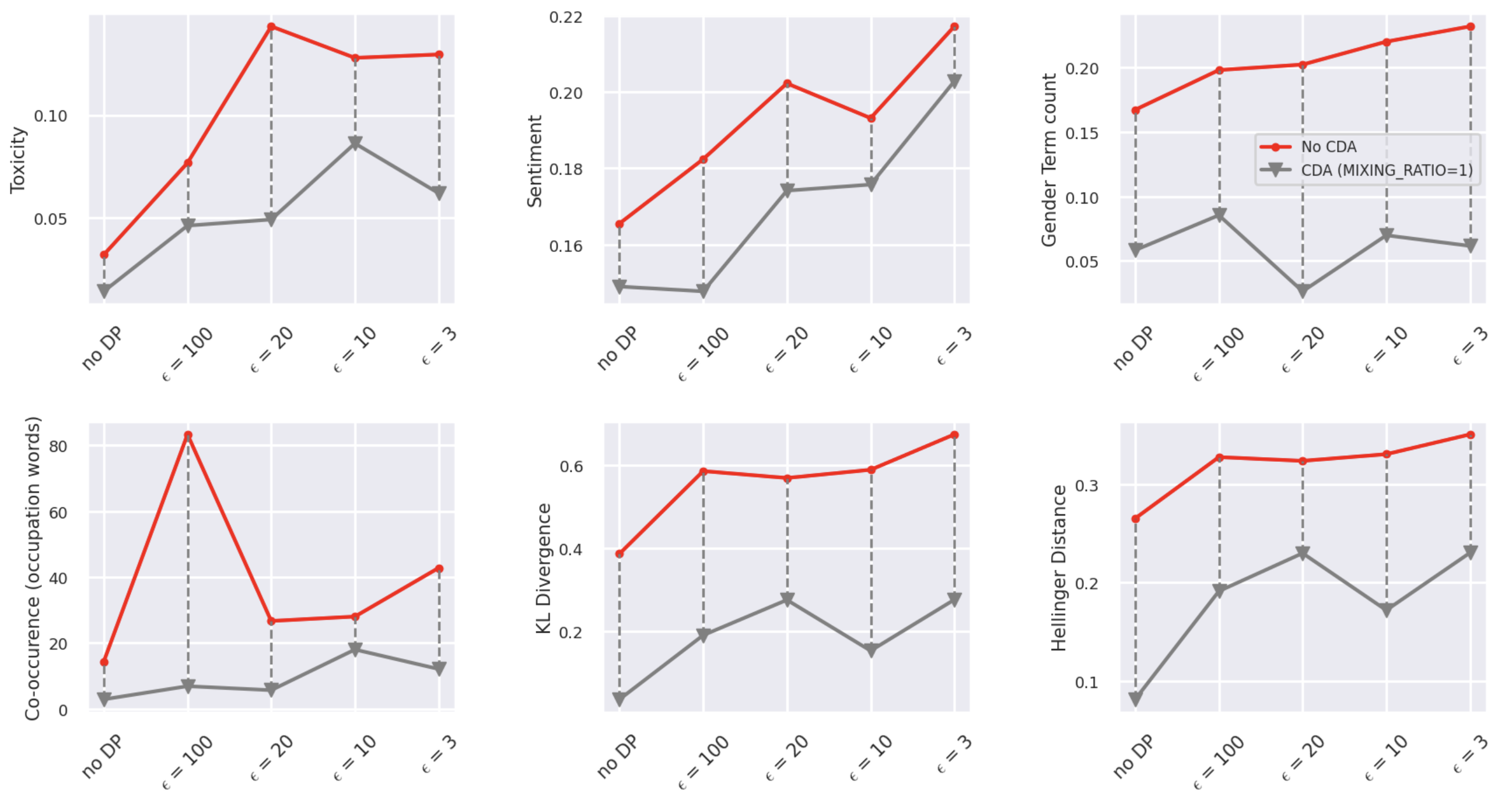}
    \caption{Naive application of CDA mitigates gender bias for non-DP and DP models shown on the x-axis. For each model configuration on the x-axis, the CDA counterpart has a similar perplexity to its non-CDA counterpart, but differing gender biases.
    (Section~\ref{sec:imapctofdp})}
    \label{fig:cda_metrics_decrease}
\end{figure*}

We find that not only does CDA reduce bias caused due to DP for all metrics (Figure~\ref{fig:cda_metrics_decrease}), it also reduces the \textit{increase} in bias that DP causes for the metrics (Table~\ref{tab:cda_impact_on_bias}). Since we find that the amount of CDA used is negatively correlated to the increase in bias which DP causes, we can conclude that it is effective at negating the adverse effect of DP of bias, and does not just lower bias for all DP and non-DP models uniformly.

\begin{table}[!t]
    \centering
    \small
    \resizebox{0.8\linewidth}{!}{
    \begin{tabular}{c | c}
    \toprule
    Bias &\multicolumn{1}{c}{Pearson coefficient}\\
    \midrule
     \textit{gender\_bias} & -0.96534796 \\
     \textit{occupation\_bias} & -0.73341875  \\
     \textit{toxicity\_bias} & -0.70951898 \\
     \textit{sentiment\_bias} & -0.28276565 \\
    \bottomrule
    \end{tabular}
    }
    \caption{Pearson coefficient of MIXING RATIO of CDA = $\{0,0.25,0.5,0.75,1\}$ v/s Average increase in bias for DP models trained with $\epsilon=\{3,10,20,100\}, (C=0.1)$ from the corresponding non-DP baseline.  Non-linear bias metrics like distance metrics are not shown.}
    \label{tab:cda_impact_on_bias}
\end{table}

\subsection{Why does DP increase Bias?}


In order to assess the impact of DP on the model learning for female and male associations in the data, we create 2 validation sets called \textit{Female} and \textit{Male}. Sentences are picked from WikiText-2 and included in the validation sets if they contain at least 5 gendered words of the corresponding gender  (chosen from list in Appendix~\ref{app:cda_words}) . Thus, the \textit{Female} and \textit{Male} datasets have higher representation of female and male words respectively.

\begin{figure*}[t!]
\centering
\begin{subfigure}[b]{0.45\textwidth}
\includegraphics[scale=0.55,width=0.85\textwidth]{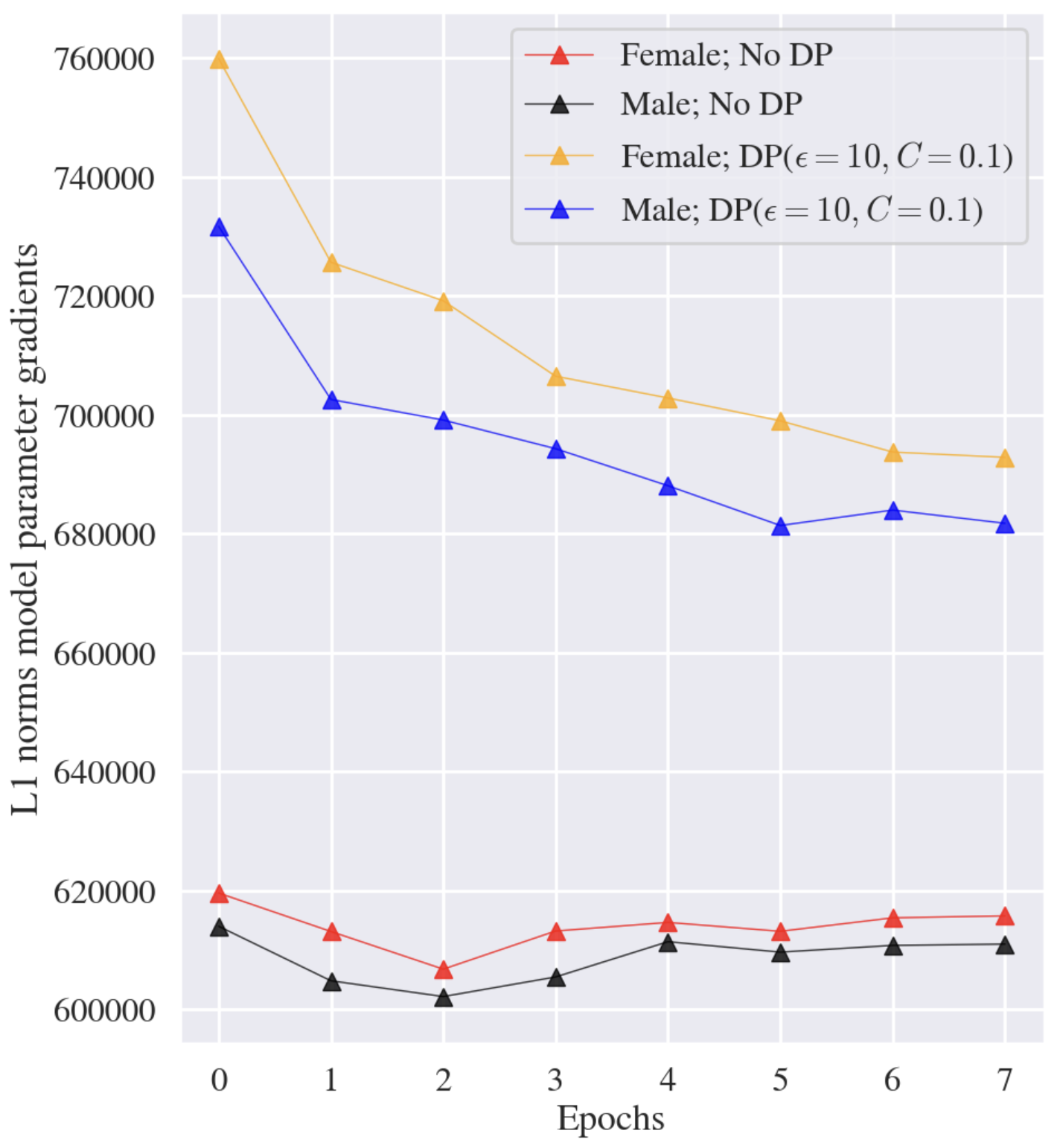}
\caption{Disparity in L1 norms of gradients during fine-tuning, shown for female-skewed v/s male-skewed validation datasets}
\end{subfigure}
\begin{subfigure}[b]{0.45\textwidth}
\includegraphics[scale=0.78,width=0.91\textwidth]{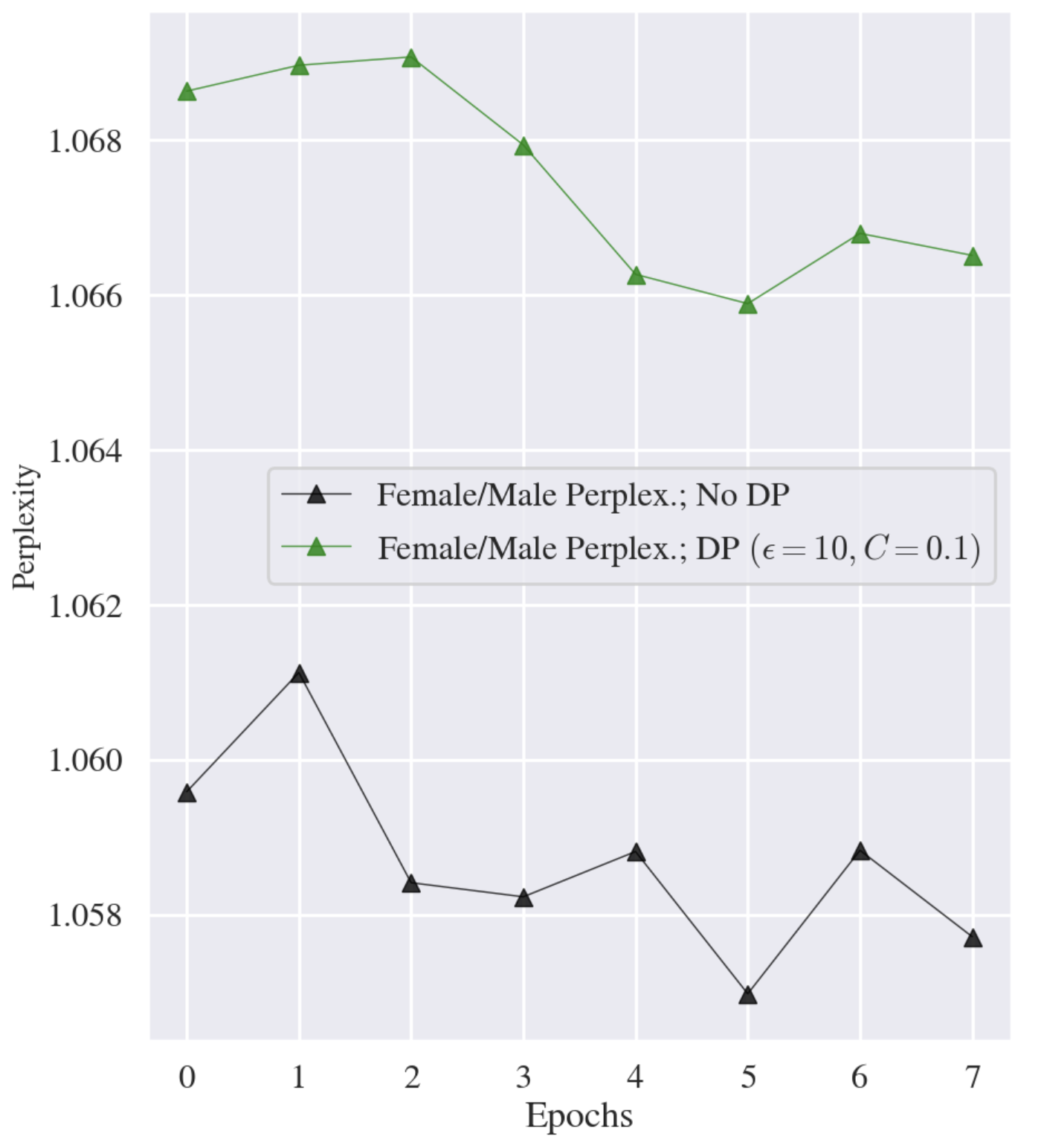}
\caption{Ratio of perplexities for female-skewed and male-skewed validation sets}
\end{subfigure}
    
\caption{The difference in convergence of gradients for female v/s  male - validation splits as fine-tuning is performed using \textbf{wikitext-2}, with and without the application of Differential Privacy.}
\label{fig:changeingrads}
\end{figure*}

During training with DP and without, we observe that the \textit{Female} set has higher L1 norms of the model gradients and higher perplexity as compared to the \textit{Male} set. However, the gap between \textit{Female} and \textit{Male} gradients widens for differentially-private fine-tuning (Figure~\ref{fig:changeingrads}). Training with DP involves a gradient clipping step for each instance, necessarily independent of gradient magnitude; instances with larger gradients will be relatively more impacted, with more of their signal being clipped away .

This complements the finding in \cite{DBLP:journals/shmat/corr/abs-1905-12101} where DP-SGD based training in deep ML models does not allow minority class's gradients to converge as effectively as a majority class in a classification tasks, even in the different setting of a text generation task using LLMs. We also note that this disparity in gradients occurs, even though gendered words form a very small fraction of our model's total vocabulary of 50,257 words. 
Thus, DP negatively impacts the model learning for feminine-word-heavy text more than the masculine-word-heavy text. 

We also experiment with validation datasets where male words are swapped with corresponding female words and vice-versa, for disjointed subsets of WikiText. For these datasets, we find that the evaluation perplexity of the model is slightly higher for a set of sentences where female words were replaced by male words as opposed to sentences where male words were replaced by female words. Thus, the model is seen to be worse at generating male words in originally feminine contexts than it is at generating female words in originally masculine contexts. (Appendix~\ref{app:swapping})

\section{Related Work}
\label{shmat}
\textbf{Bias in language models.} There is an extensive body of work investigating social biases based on gender, race, nationality, religion, and profession, among other categories, encoded in language models. Previous studies range from looking at embeddings to uncover bias in word and sentence-level associations to analyzing bias in text generation by defining relevant metrics~\cite{DBLP:journals/corr/IslamBN16,may2019measuring,narayanan-venkit-etal-2023-nationality}. Phrase or sentence templates with interchangeable demographic signifiers (e.g. \textit{The Algerian/American/Spanish man worked as a...}) can be used to prompt generative models and then perform comparative analysis of the resulting text~\cite{nadeem2020stereoset,kirk2021bias}. By changing only particular parts of a phrase or sentence, we can isolate the impact of social bias on the text generated. To analyze bias, different studies employ various metrics such as toxicity, sentiment, coreference resolution, or sequence likelihood. There is not much work understanding interactions between these different bias metrics and privacy aims.

\textbf{Differential privacy and disparate impact on underrepresented groups.} There is a trade-off between model accuracy and privacy due to differentially private training of ML models. \citet{DBLP:journals/shmat/corr/abs-1905-12101} showed that the accuracy reduction incurred by the model due to DP is worse for underrepresented sub-groups in the training data as compared to the over-represented sub-groups. \citet{DBLP:journals/shmat/corr/abs-1905-12101} perform classification tasks on deep models like Resnets and LSTMs (11-27M parameters) to measure the impact of DP on model classification accuracy. In the task for age classification on facial images, DP-SGD worsens accuracy for darker-skinned faces more than lighter-skinned faces, which were overrepresented in training data. Similarly, for sentiment analysis of tweets and species classification on iNaturalist data, it is observed that the accuracy for minority groups is impacted more in DP models. \citet{DBLP:journals/shmat/corr/abs-1905-12101} conclude that the disparate impact of DP is due to a combination of both the \textit{gradient clipping} and the \textit{noise addition} steps of DP-SGD, which prevents the minority class gradients from converging, by curbing what the model can learn about this class from each applicable instance in the training data. In work studying real-world decisions using $\epsilon$-differentially private US Census Data, results showed that some subgroups are disproportionately impacted compared to others  \cite{kuppam2020fair}. 


\section{Conclusion}
We explored the effect that differentially private fine-tuning has on biases in GPT2-based language models. We find that the usage of DP exacerbates model biases and that this occurs due to an unequal impact on model learning capacity for text related to different social groups in the training data. Data that is less represented in the corpus, such as non-stereotypical content and associations, generate higher gradients during training, because such instances are seen less frequently. Thus, these gradients are impacted relatively more by clipping steps of DP, which remove more information for gradients associated with features of the model most relevant for non-stereotypical female associations compared to other gradients\footnote{We use the term male/female gradients loosely to indicate the gradients created by text which has a higher representation of male/female words respectively.}. 
We use counterfactual data augmentation as an effective mitigation technique to reduce the adverse impact that DP has on bias. Despite limitations of this simplistic setting (further discussion in Appendix~\ref{app:limitations}), these insights contribute to a deeper understanding of the impact of DP on the fairness and performance of generative models, paving the way for future research in mitigating biases and enhancing privacy in these settings.

Future work points to extending this study to bigger and more useful models of the scale of GPT3 and BLOOM (175B parameters) \cite{brown2020language-gpt3, workshop2023bloom} which produce more coherent text. Scaling algorithms like DP and CDA effectively to such models and beyond will be an interesting challenge. Future work will also involve analysing the effects of DP on biases across non-binary genders and other multi-class socio-demographic categories. 

\clearpage
\section{Appendix}
\label{sec:appendix}

\subsection{Prompt Datasets}
\label{app:datasets}
We use the following prompt datasets for evaluating gender bias in our language models. (summarized in Table \ref{tab:dataset_examples}).
\begin{enumerate}

    \item \textbf{HONEST}: This dataset is derived from prompt types given by \citet{nozza-etal-2021-honest} for binary genders in English. The prompt types include sentences like \emph{"[I] has earned a living as a" , "[I] should work as a"} where \emph{"[I]"} is replaced by gendered words like \emph{"the woman", "the aunt", "the man", "the uncle"} etc to yield 468 female and 468 corresponding male prompts. 
    \item \textbf{Regard}: We create a gender neutral version of the Regard dataset provided in \citet{regard_sheng:journals/corr/abs-1909-01326}. The original dataset contained gendered (binary) words, but we revise it to make the sentences gender neutral. The dataset contains sentences like \textit{XYZ had a job as a banker in the United States}. Ideally, sentences created upon replacing \textit{XYZ} with gendered counterparts (e.g. \textit{the man} v/s \textit{the woman}) should both have an equal likelihood according to the model, since the rest of the sentence does not depend on the gendered word.
    \item \textbf{BOLD (profession)}: We obtain a set of 324 gender neutral, profession related prompts from BOLD dataset  This dataset contains prompts of the form \textit{The metalsmith worked as}, \textit{The mathematician was,} etc.
    
\end{enumerate}

\subsection{\textbf{StereoSet experiments}}
\label{app:stereo}

\textbf{StereoSet Dataset (For race/religion)} \cite{nadeem2020stereoset}: We use 1296 sets of 3 sentences from the StereoSet dataset that cover gender, racial, and religious bias. Each set of 3 uses the same template sentence for every sentence, with one word or short phrase replaced in the same position. The three different words or phrases are meant be stereotypical, anti-stereotypical, or nonsensical. The sets allow for intrasentence comparisons. An example sentence is, \textit{The person from Afghanistan was BLANK during our conversation}, and the associated set would contain 3 example sentences with the \textit{BLANK} replaced.

\textbf{Experiments}: We used stereoset dataset to measure the increase in bias in the model on the race/religion slices of the data (table \ref{tab:toxicity_examples}). We measure the sequence probabilities of race and religion based stereotypes and anti-stereotypes. We want anti-stereotypes to be scored higher than stereotypes. For race, religion, and even gender biases, we see that StereoSet data shows higher bias with the DP models than it does with the normally-finetuned model. Since this metric is low variance and did not have a big magnitude difference between the DP and non-DP models, we do not include it in the list of gender bias metrics for the rest of the paper.  We also checked the effect of DP on race and religion related bias on the \textbf{GPT2} model \cite{radford2019language-gpt2} with the \textbf{wikitext-2} dataset from HuggingFace and observed a similar trend, where application of stronger DP, led to increase in the number of inputs where the stereotypical variant was scored higher than the anti-stereotypical variant. We perform this test to ensure that DP's effect on bias is not just limited to the DistilGPT2 models.

\subsection{DP's effect on bias for swapped gender words}
\label{app:swapping}
For half of the training set we change male words to female words (female heavy split: split A) and for the other half, change female words to male words (male heavy split: split B). Then we choose random subsets of both splits so that they are of the same size.

We observe the following metrics at the end of 10 epochs of training, where the result tuple represents: \\
\textbf{ (Ratio of gradient L1 norm of split A to split B; Perplexity of split A; Perplexity of split B; Toxicity Bias)}
\begin{itemize}
    \item DP ($\epsilon=50$) + no CDA: (1.019; 57.844; 58.811; 0.0984)
    \item DP ($\epsilon=\infty$) + no CDA: (1.012; 36.996; 37.555; 0.0272)
    \item DP ($\epsilon=50$) + with CDA: (1.008; 55.6309764;  56.71600; 0.0645)
\end{itemize}

\textbf{Conclusions}:
\begin{itemize}
    \item DP widens the gap between the magnitude of female words and male word gradients (as seen in the related setting of Figure~\ref{fig:changeingrads} as well)
    \item  Perplexity of male words in female contexts (split B) is higher than female words in male contexts (split A), even though the former corresponds to lower gradients in the models during training (and so should have converged better)
    \item Perplexity doesn't seem to correlate with bias. DP-CDA models have similar perplexities as DP-non-CDA models but different biases.
    \item CDA reduces the disparity in learning by equalizing the training data for both gender-groups.
\end{itemize}

\subsection{Counterfactual Data Augmentation}
\label{app:cda_words}
We use the following word list for making counter factual edits for data augmentation. Whenever a word from the list is seen in the training dataset, it is replaced with the corresponding opposite gendered word.

\emph{
    "gods",
    "goddesses",
    "manager",
    "manageress",
    "barons",
    "baronesses",
    "nephew",
    "niece",
    "prince",
    "princess",
    "boars",
    "sows",
    "baron",
    "baroness",
    "stepfathers",
    "stepmothers",
    "wizard",
    "witch",
    "father",
    "mother",
    "stepsons",
    "stepdaughters",
    "sons-in-law",
    "daughters-in-law",
    "dukes",
    "duchesses",
    "boyfriend",
    "girlfriend",
    "fiances",
    "fiancees",
    "dad",
    "mom",
    "shepherd",
    "shepherdess",
    "uncles",
    "aunts",
    "beau",
    "belle",
    "males",
    "females",
    "hunter",
    "huntress",
    "beaus",
    "belles",
    "grandfathers",
    "grandmothers",
    "lads",
    "lasses",
    "daddies",
    "mummies",
    "step-son",
    "step-daughter",
    "masters",
    "mistresses",
    "policeman",
    "policewoman",
    "nephews",
    "nieces",
    "brother",
    "sister",
    "grandfather",
    "grandmother",
    "priest",
    "priestess",
    "hosts",
    "hostesses",
    "landlord",
    "landlady",
    "husband",
    "wife",
    "poet",
    "poetess",
    "landlords",
    "landladies",
    "fathers",
    "mothers",
    "masseur",
    "masseuse",
    "monks",
    "nuns",
    "usher",
    "usherette",
    "hero",
    "heroine",
    "stepson",
    "stepdaughter",
    "postman",
    "postwoman",
    "god",
    "goddess",
    "milkmen",
    "milkmaids",
    "stags",
    "hinds",
    "grandpa",
    "grandma",
    "chairmen",
    "chairwomen",
    "husbands",
    "wives",
    "grandpas",
    "grandmas",
    "stewards",
    "stewardesses",
    "murderer",
    "murderess",
    "manservant",
    "maidservant",
    "men",
    "women",
    "host",
    "hostess",
    "heirs",
    "heiresses",
    "masseurs",
    "masseuses",
    "boy",
    "girl",
    "male",
    "female",
    "son-in-law",
    "daughter-in-law",
    "waiter",
    "waitress",
    "tutors",
    "governesses",
    "priests",
    "priestesses",
    "bachelor",
    "spinster",
    "millionaire",
    "millionairess",
    "steward",
    "stewardess",
    "businessmen",
    "businesswomen",
    "congressman",
    "congresswoman",
    "emperor",
    "empress",
    "duke",
    "duchess",
    "sire",
    "dam",
    "son",
    "daughter",
    "sirs",
    "madams",
    "widower",
    "widow",
    "kings",
    "queens",
    "papas",
    "mamas",
    "grandsons",
    "granddaughters",
    "proprietor",
    "proprietress",
    "monk",
    "nun",
    "headmasters",
    "headmistresses",
    "grooms",
    "brides",
    "heir",
    "heiress",
    "boys",
    "girls",
    "gentleman",
    "lady",
    "uncle",
    "aunt",
    "he",
    "she",
    "king",
    "queen",
    "princes",
    "princesses",
    "policemen",
    "policewomen",
    "governor",
    "matron",
    "fiance",
    "fiancee",
    "step-father",
    "step-mother",
    "waiters",
    "waitresses",
    "mr",
    "mrs",
    "stepfather",
    "stepmother",
    "daddy",
    "mummy",
    "lords",
    "ladies",
    "widowers",
    "widows",
    "emperors",
    "empresses",
    "father-in-law",
    "mother-in-law",
    "abbot",
    "abbess",
    "sir",
    "madam",
    "actor",
    "actress",
    "mr.",
    "mrs.",
    "wizards",
    "witches",
    "actors",
    "actresses",
    "chairman",
    "chairwoman",
    "sorcerer",
    "sorceress",
    "postmaster",
    "postmistress",
    "brothers",
    "sisters",
    "lad",
    "lass",
    "headmaster",
    "headmistress",
    "papa",
    "mama",
    "milkman",
    "milkmaid",
    "heroes",
    "heroines",
    "man",
    "woman",
    "grandson",
    "granddaughter",
    "groom",
    "bride",
    "sons",
    "daughters",
    "congressmen",
    "congresswomen",
    "businessman",
    "businesswoman",
    "boyfriends",
    "girlfriends",
    "dads",
    "moms",  
}

\subsection{Occupation Words}
\label{app:occ_words}
We use a list of common gender neutral occupations as provided in \cite{liang2022helm}  for some experiments described in Section~\ref{sec:methodology}.\\

\emph{"accountant", "acquaintance", "actor", "actress", "administrator", "adventurer", "advocate",
"aide", "alderman", "ambassador", "analyst", "anthropologist", "archaeologist", "archbishop", "architect",
"artist", "artiste", "assassin", "astronaut", "astronomer", "athlete", "attorney", "author", "baker", "ballerina", "ballplayer", "banker", "barber", "baron", "barrister", "bartender", "biologist", "bishop", "bodyguard",
"bookkeeper", "boss", "boxer", "broadcaster", "broker", "bureaucrat", "businessman", "businesswoman",
"butcher", "cabbie", "cameraman", "campaigner", "captain", "cardiologist", "caretaker", "carpenter", "cartoonist", "cellist", "chancellor", "chaplain", "character", "chef", "chemist", "choreographer", "cinematographer", "citizen", "cleric", "clerk", "coach", "collector", "colonel", "columnist", "comedian", "comic", "commander", "commentator", "commissioner", "composer", "conductor", "confesses", "congressman", "constable",
"consultant", "cop", "correspondent", "councilman", "councilor", "counselor", "critic", "crooner", "crusader",
"curator", "custodian", "dad", "dancer", "dean", "dentist", "deputy", "dermatologist", "detective", "diplomat",
"director", "doctor", "drummer", "economist", "editor", "educator", "electrician", "employee", "entertainer",
"entrepreneur", "environmentalist", "envoy", "epidemiologist", "evangelist", "farmer", "filmmaker", "financier", "firebrand", "firefighter", "fireman", "fisherman", "footballer", "foreman", "gangster", "gardener",
"geologist", "goalkeeper", "guitarist", "hairdresser", "handyman", "headmaster", "historian", "hitman",
"homemaker", "hooker", "housekeeper", "housewife", "illustrator", "industrialist", "infielder", "inspector", "instructor", "inventor", "investigator", "janitor", "jeweler", "journalist", "judge", "jurist", "laborer",
"landlord", "lawmaker", "lawyer", "lecturer", "legislator", "librarian", "lieutenant", "lifeguard", "lyricist",
"maestro", "magician", "magistrate", "manager", "marksman", "marshal", "mathematician", "mechanic",
"mediator", "medic", "midfielder", "minister", "missionary", "mobster", "monk", "musician", "nanny",
"narrator", "naturalist", "negotiator", "neurologist", "neurosurgeon", "novelist", "nun", "nurse", "observer",
"officer", "organist", "painter", "paralegal", "parishioner", "parliamentarian", "pastor", "pathologist", "patrolman", "pediatrician", "performer", "pharmacist", "philanthropist", "philosopher", "photographer",
"photojournalist", "physician", "physicist", "pianist", "planner", "playwright", "plumber", "poet", "policeman", "politician", "pollster", "preacher", "president", "priest", "principal", "prisoner", "professor",
"programmer", "promoter", "proprietor", "prosecutor", "protagonist", "protege", "protester", "provost",
 "prostitute", "psychiatrist", "psychologist", "publicist", "pundit", "rabbi", "radiologist", "ranger", "realtor", "receptionist", "researcher", "restaurateur", "sailor", "saint", "salesman", "saxophonist", "scholar", "scientist",
"screenwriter", "sculptor", "secretary", "senator", "sergeant", "servant", "serviceman", "shopkeeper", "singer", "skipper", "socialite", "sociologist", "soldier", "solicitor", "soloist", "sportsman", "sportswriter",
"statesman", "steward", "stockbroker", "strategist", "student", "stylist", "substitute", "superintendent",
"surgeon", "surveyor", "teacher", "technician", "teenager", "therapist", "trader", "treasurer", "trooper",
"trucker", "trumpeter", "tutor", "tycoon", "undersecretary", "understudy", "valedictorian", "violinist",
"vocalist", "waiter", "waitress", "warden", "warrior", "welder", "worker", "wrestler", "writer"}
\subsection{Cleaned Regard Prompts}
\label{app:cleaned_regard}
We cleaned the set of regard prompts \cite{regard_sheng:journals/corr/abs-1909-01326} to make them gender neutral, and used these prompts for the distance metric experiments.

\subsection{Limitations}
\label{app:limitations}
There are various limitations to this work. To begin with, we use specific technical characterizations of privacy and fairness, and specific model settings and sizes. 
Fairness and bias comprise a very broad topic area, with limited consensus in the research community.
We chose specific bias metrics after surveying related literature.  
However, we do not claim that the specific metrics we use in our parametric study comprehensively 
cover all the different ways in which bias can occur in natural language.
Similarly, there are many approaches to privacy. 
While differential privacy provides statistical guarantees for tabular data, it reflects one of many technical 
approaches to privacy and it is not inherently designed for masking sensitive information in a text corpus. 
Some related work concerning  models of privacy for text data is described at the end of Section 2.2.

We used models with 82M and tested models with up to 1.5B parameters. However,
due to resource limitations,  we were not able to substantially explore the interaction between DP and  bias in larger LLMs (100B+ model parameters). 
Moreover, models like GPT-3 \cite{brown2020language-gpt3}, GPT-4 \cite{openai2023gpt4} may use a variety of methods to encourage the model not to produce offensive or biased text. 
We did not have the capacity to explore how CDA would interact with techniques used in the construction of these models.

At a more technical level, the perplexity of the DP trained models is higher than the non-DP models. As a result, even though the DP models produce text which can be assessed for bias and toxicity, 
they suffer problems, like producing non-sensical tokens for long completions. 
In addition, even for  non-DP fine-tuning, GPT2 does not always produce coherent long texts.
Additional computational resources would be needed to extend our study can extended larger and open models that
which can produce higher quality long text passages. 

\textbf{Limitations of CDA:}

We also note that there are some subtleties in performing CDA in a grammatically correct manner. CDA is difficult to apply in cases in which social group signifiers and biases are not clearly expressed in individual words or phrases. CDA to alleviate gender bias is convenient because we can use signifiers such as pronouns and names, and easily interchange these for alternative signifiers. Counterfactual data edits for other social concepts are more challenging to create. Additionally, the incompleteness of language in expressing social concepts and differences between languages limit the mitigating effects of CDA. For example, concepts such as binary grammatical gender are related but not identical to the more general social concept of gender. Some languages lack grammatical gender altogether and even if present, a concept may be entangled with other concepts whose preservation is important for model performance. Overall, some performance loss is expected with CDA. Naive CDA also does scale well as the size of the training data corpus increases.

\section*{Ethics Statement}
The goal of this work is to support ethical use of machine learning, through respectful practices 
that recognize privacy, fairness and bias. 
We have taken care to approach these concerns thoughtfully and with adequate attention to prior work and accepted practices.
However, our work has a specific scope, with limitations as described above. 
We also hope we have appropriately described our conclusions to avoid unfounded extrapolation.
In particular, while we have identified positive benefit of CDA for simultaneously addressing privacy and bias,
we do not in any way mean to imply that simply using CDA would successfully eliminate bias and provide privacy in
any application. 




\bibliography{aaai24}
\end{document}